\documentclass{bmvc2k}

\usepackage{amsmath, amssymb, booktabs, mathrsfs, multirow, float, enumitem}

\title{SVAC: Scaling Is All You Need For Referring Video Object Segmentation}

\addauthor{Li Zhang}{lz2714@columbia.edu}{1}
\addauthor{Haoxiang Gao}{haoxiang@alumni.cmu.edu}{2}
\addauthor{Zhihao Zhang}{z.zhihao@columbia.edu}{1}
\addauthor{Luoxiao Huang}{lh2583@nyu.edu}{3}
\addauthor{Tao Zhang}{zhang_tao@whu.edu.cn}{4*}

\addinstitution{
 Columbia University\\
 New York, USA
}
\addinstitution{
 Carnegie Mellon University\\
 Pittsburgh, USA
}
\addinstitution{
 New York University\\
 New York, USA
}
\addinstitution{
 Wuhan University \\
 Wuhan, China
}

\runninghead{Li Zhang}{SVAC}

\begin{document}

\maketitle

\begin{abstract}
Referring Video Object Segmentation (RVOS) aims to segment target objects in video sequences based on natural language descriptions. While recent advances in Multi-modal Large Language Models (MLLMs) have improved RVOS performance through enhanced text-video understanding, several challenges remain, including insufficient exploitation of MLLMs’ prior knowledge, prohibitive computational and memory costs for long-duration videos, and inadequate handling of complex temporal dynamics. In this work, we propose \textbf{SVAC}, a unified model that improves RVOS by scaling up input frames and segmentation tokens to enhance video-language interaction and segmentation precision. To address the resulting computational challenges, \textbf{SVAC} incorporates the \textbf{Anchor-Based Spatio-Temporal Compression (ASTC)} module to compress visual tokens while preserving essential spatio-temporal structure. Moreover, the \textbf{Clip-Specific Allocation (CSA)} strategy is introduced to better handle dynamic object behaviors across video clips. Experimental results demonstrate that \textbf{SVAC} achieves state-of-the-art performance on multiple RVOS benchmarks with competitive efficiency. Our code is available at \url{https://github.com/lizhang1998/SVAC}.

\end{abstract}

\section{Introduction}
\label{sec:intro}

Referring Video Object Segmentation (RVOS)~\cite{MeViS, Gavrilyuk2018ActorAA, khoreva2019video, seo2020urvos} is a rapidly evolving task that aims to segment target objects in a video sequence based on natural language descriptions. Traditional RVOS methods~\cite{MeViS, yu2018mattnet, GRES, khoreva2019video, zhang2023dvis, zhang2025dvis++, zhou2024improving}  focus on designing effective fusion strategies for text and video features to align textual queries with visual content. Recent advancements~\cite{Lai_2024_CVPR, yan2024visa, sa2va, bai2024tokensegalllanguage} leveraging Multi-modal Large Language Models (MLLMs) have revolutionized RVOS, achieving superior performance through enhanced understanding and reasoning capabilities over text and video interactions. By integrating multi-modal inputs into a unified framework, MLLMs enable robust alignment of dynamic visual content with complex textual descriptions, setting new benchmarks in RVOS.

\begin{figure}
\centering

\begin{subfigure}{}
\centering
\includegraphics[width=0.9\textwidth]{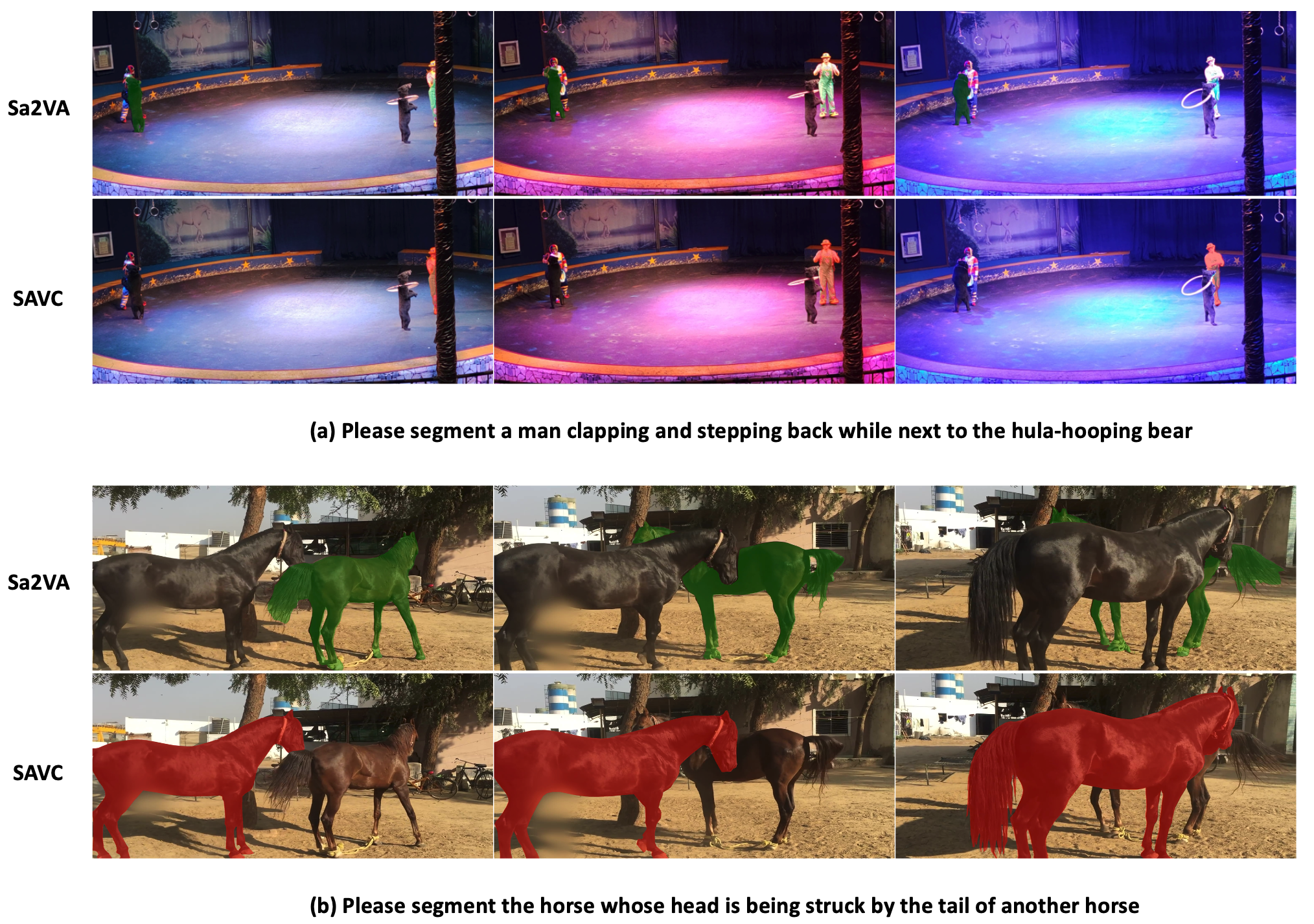}
\caption{\textbf{Visualization of Comparison between Sa2VA and our SVAC.} (a) Sa2VA shows poor contextual comprehension by segmenting a bear when explicitly instructed to segment a person, while our SVAC correctly localizes and segments the human subject as specified. (b)When confronted with complex motion patterns, such as two horses occluding each other and moving rapidly, Sa2VA’s segmentation capabilities deteriorate significantly, whereas SVAC maintains accurate segmentation performance.}
\label{fig:demo}
\end{subfigure}

\vspace{2mm}  

\begin{subfigure}{}
\centering
\includegraphics[width=0.9\textwidth]{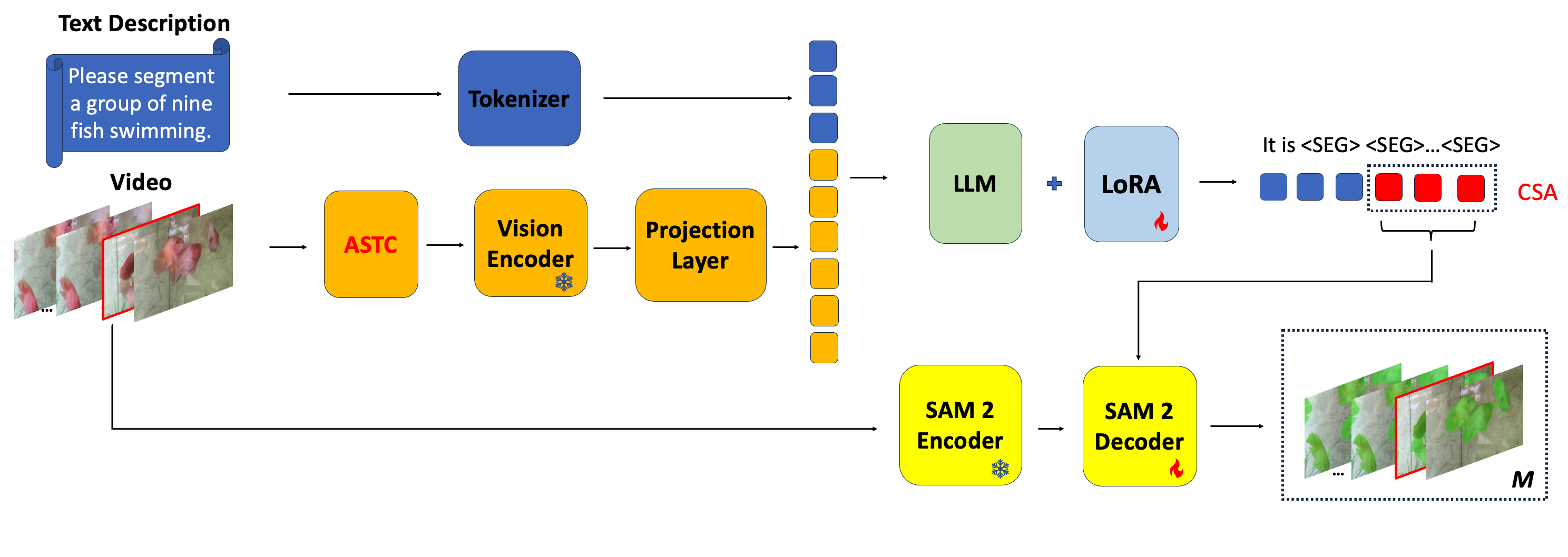}
\caption{\textbf{Overall Structure.} The pipeline consists of three main stages: (1) Text and video frames are encoded and projected into a shared embedding space before being fed into the LLM; (2) <SEG> tokens are extracted from the LLM output; (3) The SAM-2 Decoder generates the final mask by combining features from the SAM-2 Encoder with the extracted <SEG> tokens.}
\label{fig:struc}
\end{subfigure}
\end{figure}

Despite the remarkable progress driven by MLLM-based models like Sa2VA~\cite{sa2va} and its predecessors, several critical challenges persist, limiting the scalability and robustness of RVOS. These include: \textbf{(1) Insufficient exploitation of MLLMs’ prior knowledge}: Existing approaches~\cite{Lai_2024_CVPR, yan2024visa, sa2va}  often under-utilize the rich prior knowledge within MLLM due to feeding it only a sparse selection of frames. The resulting limited spatial and temporal information hinders the MLLM's ability to accurately locate objects referenced in the text descriptions, as shown in Fig.~\ref{fig:demo} (a);  \textbf{(2) Prohibitive computational cost and memory usage for high-resolution, long-duration videos}: Processing high-resolution video inputs places a significant burden on MLLM architectures, especially Transformer-based models with their quadratic attention complexity. Furthermore, maintaining temporal coherence across extended video sequences necessitates the costly storage and updating of extensive feature maps or hidden states, resulting in substantial memory overhead; \textbf{(3) Inadequate handling of complex temporal dynamics}: Current methods~\cite{Lai_2024_CVPR, yan2024visa, sa2va, bai2024tokensegalllanguage}  rely on a single segmentation token to represent object localization throughout the entire video. However, for objects in rapid motion, a single token is insufficient to capture dynamic changes, as it cannot adapt to varying object positions and appearances across frames (see in Fig.~\ref{fig:demo} (b)).

In this work, regarding \textit{challenge (1)}, we observe that scaling up video frames significantly enhances MLLM performance in RVOS. The previous SOTA model, Sa2VA, only samples 5 frames from the whole video, often failing to segment objects in the middle portion of the video. By scaling up to 10 or 20 frames by providing richer spatial and temporal cues, this issue is alleviated. However, as noted in \textit{challenge (2)}, further increasing the frame number substantially increases computational and memory demands.


To overcome \textit{challenge (2)}, the key lies in reducing the number of visual tokens before input to LLM. We propose a novel compression method called \textbf{Anchor-Based Spatio-Temporal Compression} (\textbf{ASTC} in Section~\ref{sec:astc}) that optimizes the trade-off between processing a larger number of frames and preserving fine-grained details critical for accurate segmentation. Unlike existing pooling methods~\cite{weng2024longvlmefficientlongvideo, li2024llamavid}, which often compromise on spatial resolution or temporal coherence, and in contrast to merging strategies~\cite{yang2024visionzip, li2024tokenpackerefficientvisualprojector} that necessitate structural modifications to the MLLM, ASTC method divides videos into clips, retaining the first frame and compressing remaining frames into a single resized composite, preserving critical spatial-temporal features. It achieves superior memory efficiency, reduced computational complexity, and enhanced segmentation accuracy without requiring any change to the model structure, ensuring seamless plug-in compatibility.


Finally, to address the \textit{challenge (3)} of complex object motion in videos, we scale up the number of segmentation tokens with the \textbf{Clip-Specific Allocation} (\textbf{CSA} in Section~\ref{sec:seg}) strategy. Previous approaches~\cite{bai2024tokensegalllanguage, Lai_2024_CVPR, yan2024visa, sa2va} typically use a single segmentation token per object through the entire video, which struggles with nuanced temporal dynamics such as rapid trajectory shifts or occlusions. Our method addresses this by assigning a dedicated segmentation token to each video clip. This refined strategy significantly enhances segmentation accuracy by more effectively capturing intricate motion patterns within these shorter temporal segments.

Based on the above discussions, we present \textbf{SVAC}: \textbf{S}caling up both \textbf{V}ideo frames with the \textbf{A}STC method and segmentation tokens with the \textbf{C}SA strategy. Built upon the Sa2VA framework, SVAC achieves SOTA performance on multiple RVOS benchmarks. Overall, our main contributions are summarized as follows:
\begin{itemize}[itemsep=0em, parsep=0pt, topsep=0.2em]
    \item We discover that scaling up video frames input to MLLM significantly boosts its performance in RVOS.
    \item To mitigate the increased computational and memory demands, we propose Anchor-Based Spatio-Temporal Compression (ASTC, see Section~\ref{sec:astc}) method to reduce visual tokens prior to being input to LLM. Additionally, to handle complex object motion in videos, we scale up segmentation tokens with Clip-Specific Allocation (CSA in Section~\ref{sec:seg}) strategy, rather than leveraging a single token for the entire video.
    \item We develop a unified model named SVAC, and extensive experiments show that it achieves SOTA performance on multiple RVOS benchmarks.
\end{itemize}

\section{Related Work}
\label{sec:relat}
{\bf Multi-modal Large Language Models.} The development of Multimodal Large Language Models (MLLMs) has accelerated rapidly in recent years, enabling unified understanding and reasoning across vision and language modalities. Early efforts~\cite{radford2021learning, jia2021scaling, yao2022filip, yuan2021florencenewfoundationmodel} such as CLIP~\cite{radford2021learning} establish the foundation by learning joint image-text representations through contrastive learning. This is followed by models like Flamingo~\cite{alayrac2022flamingo}, which fuses vision inputs into frozen large language models (LLMs) using lightweight cross-attention modules.
A key trend in MLLM design is the decoupling of vision encoders and language backbones. Vision modules such as ViT~\cite{dosovitskiy2020vit}, EVA~\cite{fang2023eva}, or CLIP encoders are often paired with powerful LLMs, including LLaMA~\cite{touvron2023llamaopenefficientfoundation}, ChatGLM~\cite{glm2024chatglmfamilylargelanguage}, Qwen~\cite{yang2024qwen2technicalreport}, Vicuña~\cite{vicuna2023} and InternLM~\cite{cai2024internlm2}. These LLMs provide strong language understanding and generation capabilities, and are typically aligned with vision features through learned projection layers or Q-formers, as demonstrated in BLIP-2~\cite{li2023blip}, LLaVA~\cite{liu2023llava} and others~\cite{zhu2023minigpt, chen2024internvl, bai2023qwenvlversatilevisionlanguagemodel, wang2024qwen2vlenhancingvisionlanguagemodels, chen2025expandingperformanceboundariesopensource, wang2024world, wang2025vgr, lei2025scalability, xiao2024seeing, wang2025traceable, li2025denseworld}. Among these MLLMs, we choose InternVL2-5~\cite{chen2025expandingperformanceboundariesopensource, dong2024benchmarking} as our backbone due to its superior performance in multimodal tasks such as VQA, image captioning, and document understanding.
\vspace{10pt}
\\
{\bf Referring Video Object Segmentation.} Referring Video Object Segmentation (RVOS) seeks to segment a target object throughout a video based on a natural language expression. Early approaches~\cite{MeViS, yu2018mattnet, GRES, khoreva2019video, zhang2023dvis, zhang2025dvis++, zhou2024improving, niu2025beyond} explore various fusion modules to integrate visual and linguistic features, achieving improved performance in generating accurate masks. Subsequent works~\cite{UNINEXT} leverage transformer-based architectures to unify segmentation and tracking in video, enabling more robust handling of temporal dynamics. With the advent of MLLM, many works~\cite{Lai_2024_CVPR, zhang2024omg, hanoona2023GLaMM, sa2va, zhang2025pixel} attempt to introduce MLLM into the field of referring segmentation. For instance, LISA~\cite{Lai_2024_CVPR} proposes reasoning-based segmentation, while GLaMM~\cite{hanoona2023GLaMM} curates a new dataset to support region-level captioning and segmentation tasks. Similarly, VISA~\cite{yan2024visa} enhances segmentation by selecting textrelevant frames for MLLM processing. More recently, Sa2VA~\cite{sa2va} integrates SAM-2~\cite{ravi2024sam2} with MLLM, achieving strong performance across both image and video referring segmentation tasks, as well as visual question answering (VQA). However, existing methods are limited by processing only a small number of frames and relying on a single segmentation token, which constrains the capacity of MLLMs and hinders the modeling of fine-grained temporal dynamics. In contrast, our proposed SVAC model employs ASTC to efficiently process substantially more frames, fully harnessing the potential of MLLMs. Additionally, by scaling segmentation tokens through CSA, our method captures subtle temporal variations with higher precision.
\vspace{10pt}
\\
{\bf Compression.} Compression techniques for Vision Transformers (ViTs) and Video-Language
Models (VLMs) aim to reduce token redundancy while preserving performance. A notable approach, Token Merging (ToMe)~\cite{bolya2023tokenmergingvitfaster}, employs Bipartite Soft Matching to merge similar tokens across ViT layers. However,
it often loses fine-grained details essential for tasks like Referring Video
Object Segmentation (RVOS) and requires intrusive model modifications,
reducing its flexibility. Subsequent works like VisionZip~\cite{yang2024visionzip} and TokenPacker~\cite{li2024tokenpackerefficientvisualprojector} explore alternative similarity metrics but face similar limitations in detail-sensitive applications.
In contrast, methods like LongVLM~\cite{weng2024longvlmefficientlongvideo} and LLaMA-VID~\cite{li2024llamavid} use
average pooling on vision tokens post-encoding to achieve compression.
While more adaptable, these approaches still struggle to retain critical
details, making them less effective for RVOS.
To address these challenges, we propose ASTC method, which ensures detail preservation in RVOS 
while maintaining computational efficiency.

\section{Method}
\label{sec:method}
\subsection{Task Description}
In this section, we would describe RVOS task more specifically. Formally, given a video sequence $V = \{I_1, I_2, \dots, I_T\}$, where each frame $I_t \in \mathbb{R}^{H \times W \times 3}$ represents an RGB image of spatial resolution $H \times W$ and temporal length $T$, and a referring expression $S$ composed of $L$ words, i.e., $S = \{w_1, w_2, \dots, w_L\}$, the goal is to predict a sequence of binary segmentation masks $M = \{M_1, M_2, \dots, M_T\}$, where each $M_t \in \{0, 1\}^{H \times W}$ corresponds to the pixel-wise mask of the referred object in frame $I_t$.

The task can be formulated as learning a mapping function:
\begin{equation}
f: (V, S) \rightarrow M
\end{equation}

where the model $f$ learns to align the spatial-temporal visual features of the video $V$ with the semantic representation of the referring expression $S$, to produce frame-level object masks that are consistent with both the visual context and the language description.

\subsection{Overall Architecture}
The overall structure is shown in Fig.~\ref{fig:struc}.
\\
\textbf{Pretrained MLLM.} The core of our structure is a pretrained MLLM, which consists of three key components: a vision encoder, a projection layer, and a large language model (LLM). The vision encoder processes input frames, compressed by ASTC model, to extract compact visual features. These features are then mapped by the projection layer into visual tokens that align with the text token embedding space, enabling seamless integration with text tokens. The combined sequence of visual and text tokens is fed into the LLM, which generates predictions by modeling the joint distribution of multimodal inputs.
\\
\textbf{Segmentation.} To enable precise video segmentation, the system integrates SAM-2 with the MLLM using a list of \texttt{<SEG>} token as a bridge. The MLLM processes the input sequence and extracts the hidden states of the \texttt{<SEG>} tokens, which serve as a spatial-temporal prompt for the SAM-2 Decoder. This prompt provides contextual cues from the multimodal input to guide segmentation. Meanwhile, the SAM-2 Encoder extracts frame-level features from the input video frames and feeds them to the Decoder. By combining the MLLM’s prompt with these frame features, the SAM-2 Decoder generates accurate, temporally consistent segmentation masks across the video.

\begin{figure}
\centering
\includegraphics[width=0.9\textwidth]{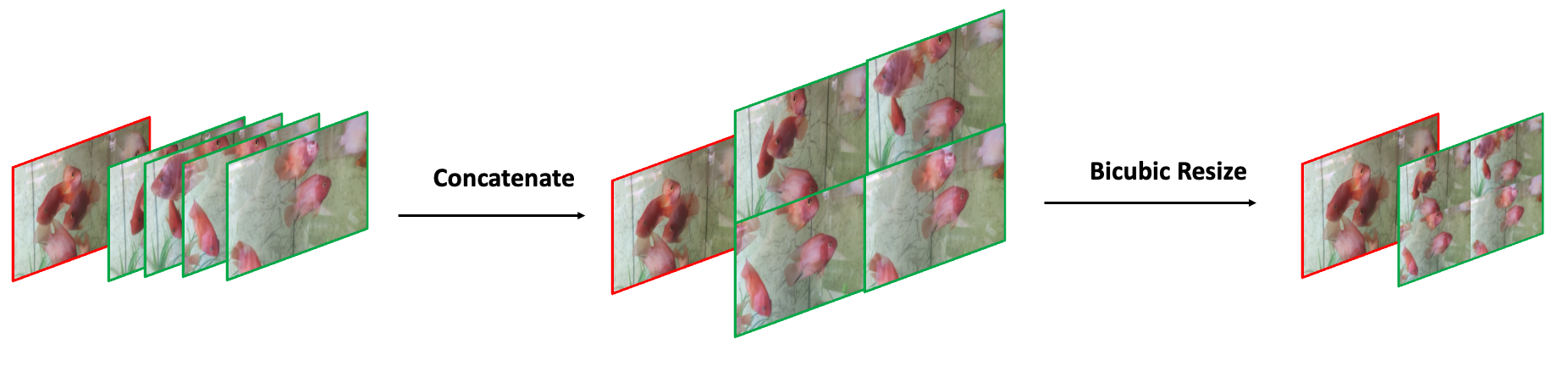}
\caption{\textbf{Anchor-Based Spatio-Temporal Compression (ASTC).} For a 5-frame video clip, we preserve the first frame as anchor while compressing the remaining 4 frames through temporal concatenation (left-to-right, top-to-bottom arrangement) and subsequent resizing to match the anchor frame dimensions.}
\label{fig:ASTC}
\end{figure}

\subsection{Scaling Frames}
\label{sec:astc}
During testing with the Sa2VA model, we observe that increasing the number of sampled frames per video significantly improves performance. Initially, Sa2VA samples only five frames, but scaling to more frames causes out-of-memory issues. To mitigate the computational burden of processing dense visual tokens while preserving critical spatio-temporal information, we introduce \textbf{Anchor-Based Spatio-Temporal Compression} (\textbf{ASTC}), as shown in Fig.~\ref{fig:ASTC}.

Given a video sequence \( V = \{I_1, I_2, \dots, I_T\} \), where each frame \( I_t \in \mathbb{R}^{H \times W \times 3} \) represents an RGB image at time step \( t \), we first partition the sequence into \( N \) non-overlapping clips of equal temporal length \( m \). Each clip is denoted as \( C^{(i)} = \{I_1^{(i)}, I_2^{(i)}, \dots, I_m^{(i)}\} \) for \( i = 1, \dots, N \). Within each clip \( C^{(i)} \), the first frame \( I_1^{(i)} \) is designated as the anchor frame and retained at its full spatial resolution to anchor the spatial context. The subsequent frames \( \{I_2^{(i)}, I_3^{(i)}, \dots, I_m^{(i)}\} \), which capture the temporal evolution, undergo a dual compression process in both spatial and temporal dimensions.

First, the subsequent frames are concatenated along the spatial dimensions while preserving the channel dimension. The concatenation is performed in temporal order, arranging frames from left to right and top to bottom to preserve temporal coherence. To maintain the aspect ratio of the original frames \( H \times W \), these \( m-1 \) frames are arranged into a 2D grid layout, forming an aggregated tensor:

\begin{equation}
I_{\mathit{agg}}^{(i)} = \mathit{Concat}(I_2^{(i)}, I_3^{(i)}, \dots, I_m^{(i)}) \in \mathbb{R}^{H' \times W' \times 3}
\label{eq:eq1}
\end{equation}

Here, as defined in Equation~\ref{eq:eq1}, \( H' \) and \( W' \) are chosen such that the grid layout accommodates all \( m-1 \) frames while preserving an aspect ratio close to \( H \times W \). If the spatial dimensions \( H' \times W' \) slightly exceed what is required to exactly fit the frames, the remaining unused regions in the grid are filled with zeros (zero-padding) to maintain uniformity.

Next, applying bicubic interpolation to resize \( I_{\text{agg}}^{(i)} \) back to the original spatial dimensions of a single frame:

\begin{equation}
\tilde{I}_{\mathit{agg}}^{(i)} = \mathit{Bicubic\_Resize}(I_{\mathit{agg}}^{(i)}, H \times W \times 3) \in \mathbb{R}^{H \times W \times 3}
\end{equation}

This compressed representation \( \tilde{I}_{\text{agg}}^{(i)} \) effectively encapsulates the temporal dynamics of frames \( \{I_2^{(i)}, \dots, I_m^{(i)}\} \) while reducing the spatial and channel footprint.

The final representation of each clip \( C^{(i)} \) is the pair: \( \{I_1^{(i)}, \tilde{I}_{\text{agg}}^{(i)}\} \), where \( I_1^{(i)} \) serves as the anchor frame with full spatial fidelity, and \( \tilde{I}_{\text{agg}}^{(i)} \) provides a compressed embedding of the temporal dynamics across the remaining frames.

Assuming a vision encoder generates \( s \) tokens per frame, the token count for the original clip is \( s_{\mathit{original}} = m \cdot s \). After applying ASTC, the token count is reduced to \( s_{\mathit{reduced}} = 2 \cdot s \), yielding a compression ratio of   \( 2 / m \). This approach significantly enhances computational efficiency and memory utilization while maintaining both the spatial detail of the anchor frame and the temporal context of the clip.
\subsection{Scaling Tokens}
\label{sec:seg}
In order to enhance the model's ability to distinguish between different temporal segments, we introduce \textbf{Clip-Specific Allocation} (\textbf{CSA}) to scale up the number of $\texttt{<SEG>}$ tokens: Given a video that has been divided into $N$ clips $\{C^{(1)}, C^{(2)}, ..., C^{(N)}\}$, we assign a unique segment token $\texttt{<SEG>}^{(i)}$ to each clip $C^{(i)}$. 

Formally, let $S = \{\texttt{<SEG>}^{(1)}, \texttt{<SEG>}^{(2)}, ..., \texttt{<SEG>}^{(N)}\}$ be the set of clip-specifc segmentation tokens, where each $\texttt{<SEG>}^{(i)} \in \mathbb{R}^d$ and $d$ is the embedding dimension. Instead of using a single, shared $\texttt{<SEG>}$ token for all frames as a global prompt, we concatenate all the segmentation tokens and input them to SAM-2. For the frame \( I_{t}^{(i)} \), which denotes the frame at time step \( t \) within clip $C^{(i)}$, we compute the mask \( M_{t} \) as follows:

\begin{equation}
M_{t} = \mathit{SAM2\_Decoder}\big(\mathit{SAM2\_Encoder}(I_{t}^{(i)}), \ \texttt{<SEG>}^{(i)})
\end{equation}

By providing temporally-aware $\texttt{<SEG>}$ tokens, the model can better localize objects and segment changes over time, as each $\texttt{<SEG>}$ token encodes clip-specific temporal priors and scene context. 

\begin{table}[h]
\centering
\caption{Referring Video Object Segmentation Performance}
\footnotesize
\begin{tabular}{l|c|ccc|ccc|ccc}
\toprule
\multirow{2}{*}{Method} & \multirow{2}{*}{Year} & \multicolumn{3}{c|}{MeVIS$_{val\_u}$~\cite{MeViS}} & \multicolumn{3}{c|}{Ref-DAVIS17~\cite{khoreva2019video}} & \multicolumn{3}{c}{ReVOS~\cite{yan2024visa}} \\
\cmidrule(lr){3-5} \cmidrule(lr){6-8} \cmidrule(lr){9-11}
 & & $J$ & $F$ & $J\&F$ & $J$ & $F$ & $J\&F$ & $J$ & $F$ & $J\&F$ \\
\midrule
URVOS~\cite{seo2020urvos} & ECCV 2020 & - & - & - & 47.3 & 56.0 & 51.6 & - & - & - \\
MTTR~\cite{botach2021end} & CVPR 2022 & - & - & - & - & - & - & 25.1 & 25.9 & 25.5 \\
ReferFormer~\cite{wu2022referformer} & CVPR 2022 & - & - & - & 58.1 & 64.1 & 61.1 & 26.2 & 29.9 & 28.1 \\
LISA-13B~\cite{Lai_2024_CVPR} & CVPR 2024 & - & - & - & 63.2 & 68.8 & 66.0 & 39.8 & 43.5 & 41.6 \\
TrackGPT-13B~\cite{stroh2024trackgptgenerativepretrained} & ArXiv2024 & - & - & - & 62.7 & 70.4 & 66.5 & 43.2 & 46.8 & 45.0 \\
VISA-13B~\cite{yan2024visa} & ECCV 2024 & - & - & - & 67.0 & 73.8 & 70.4 & 48.8 & 52.9 & 50.9 \\
VideoLISA~\cite{bai2024tokensegalllanguage} & NeurIPS 2024 & 50.9 & 58.1 & 54.5 & 72.7 & 64.9 & 68.8 & - & - & - \\
VideoGLaMM~\cite{munasinghe2024videoglamm} & CVPR 2025 & - & - & - & 73.3 & 65.6 & 69.5 & - & - & - \\
GLUS~\cite{lin2025glus} & CVPR 2025 & - & - & 61.6 & - & - & - & 47.5 & 52.4 & 49.9 \\
ViLLa~\cite{zheng2025villavideoreasoningsegmentation} & ArXiV 2025 & - & - & - & 70.6 & 78.0 & 74.3 & 54.9 & 59.1 & 57.0 \\
\midrule
Sa2VA-1B~\cite{sa2va} & ArXiV 2025 & - & - & 53.4 & - & - & 69.5 & - & - & 47.6\\
Sa2VA-4B~\cite{sa2va} & ArXiV 2025 & - & - & 55.9 & - & - & 73.7 & - & - & 53.2\\
Sa2VA-8B~\cite{sa2va} & ArXiV 2025 & - & - & 58.9 & - & - & 75.9 & - & - & 57.6\\
Sa2VA-26B~\cite{sa2va} & ArXiV 2025 & - & - & 61.8 & - & - & 78.6 & - & - & 58.4\\
\midrule
SVAC-1B (Ours) & BMVC 2025 & 55.8 & 64.6 & 60.2 & 67.5 & 76.1 & 71.8 & 48.1 & 50.7 & 49.4\\
SVAC-4B (Ours) & BMVC 2025 & 56.9 & 65.6 & 61.3 & 71.4 & 80.0 & 75.7 & 55.1 & 56.1 & 55.6\\
SVAC-8B (Ours) & BMVC 2025 & \textbf{58.2} & \textbf{67.3} & \textbf{62.8} & \textbf{74.9} & \textbf{83.8} & \textbf{79.4} & \textbf{58.2} & \textbf{60.2} & \textbf{59.2}\\
\bottomrule
\end{tabular}
\label{tab:results}
\end{table}

\section{Experiments}
\label{sec:exp}
\subsection{Experimental Settings}
{\bf BaseLine.} We construct the baseline by integrating the
SOTA MLLM models like InternVL2-5~\cite{chen2025expandingperformanceboundariesopensource} and SAM-2~\cite{ravi2024sam2}, following the Sa2VA~\cite{sa2va} framework. We sample 100 frames per video and apply the ASTC method before feeding them to the MLLM. In ASTC, frames are divided into 10 clips, and clip-specific segmentation allocation generates 10 \texttt{<SEG>} tokens, whose hidden states are processed by the SAM-2 decoder to produce the final mask.
\\
{\bf Dataset \& Metrics.} We fine-tune on three major 
RVOS datasets: MeVIS~\cite{MeViS} dataset,  Ref-YoutubeVOS~\cite{seo2020urvos} dataset, and ReVOS~\cite{yan2024visa} dataset. Performance is evaluated using the $J$(average IoU), $F$ (boundary $F$ measure), and $J\&F$(average of $J$ and $F$) metric.
\\
{\bf Implementation Details.} Training and testing are conducted using the XTuner codebase. The initial learning rate is set to 4e-5, with the perception model frozen and the LLM fine-tuned using LoRA. The LLM's maximum sequence length is 12,288. Each video is sampled at 100 frames, grouped into clips of 10 frames each. All experiments are performed on 8 NVIDIA A800 GPUs with 40GB of memory.
\subsection{Main Results}
Table~\ref{tab:results} demonstrates the SOTA performance of our SVAC model across multiple benchmarks. With only 8B parameters, SVAC achieves SOTA $J\&F$ scores of 62.8, 79.4, and 59.2 on MeVIS(val\_u)~\cite{MeViS}, Ref-DAVIS17~\cite{khoreva2019video}, and ReVOS~\cite{yan2024visa} datasets respectively, surpassing the previous SOTA model Sa2VA-26B by margins of 1.0, 0.8, and 0.8 points.
\subsection{Ablation Study}
To better understand the impact of each component in our framework, we conduct a series of ablation studies. Unless otherwise specified, all experiments are performed on the MeVIS dataset using the 1B variants of the MLLM backbone.
\vspace{10pt}

\noindent\textbf{Compression Method.}
We compare our proposed ASTC method againt several common compression techniques, including pooling~\cite{weng2024longvlmefficientlongvideo, li2024llamavid}, token pruning~\cite{tan2025tokencarveinformationpreservingvisualtoken}, token merging~\cite{yang2024visionzip, bolya2023tokenmergingvitfaster, li2024tokenpackerefficientvisualprojector}. For all methods, we set a compression ratio of 25\% and sample 80 frames. The evaluated methods are: (1) Average Pooling with $2\times2$ kernel and stride of 2; (2) Max Pooling with $2\times2$ kernel and stride of 2; (3) Token Pruning, retaining only the top 25\% of tokens based on attention scores to the \texttt{<CLS>} token; (4) Token Merging, uniformly sampling 25\% of tokens and merging the remaining tokens into them; (5) ASTC, configured with 8 frames per clip. As shown in Table~\ref{tab:compression}, ASTC outperforms all other methods by at least 1 point.
\vspace{5pt}


\begin{figure}[htbp]
    \centering
    \includegraphics[width=0.8\linewidth]{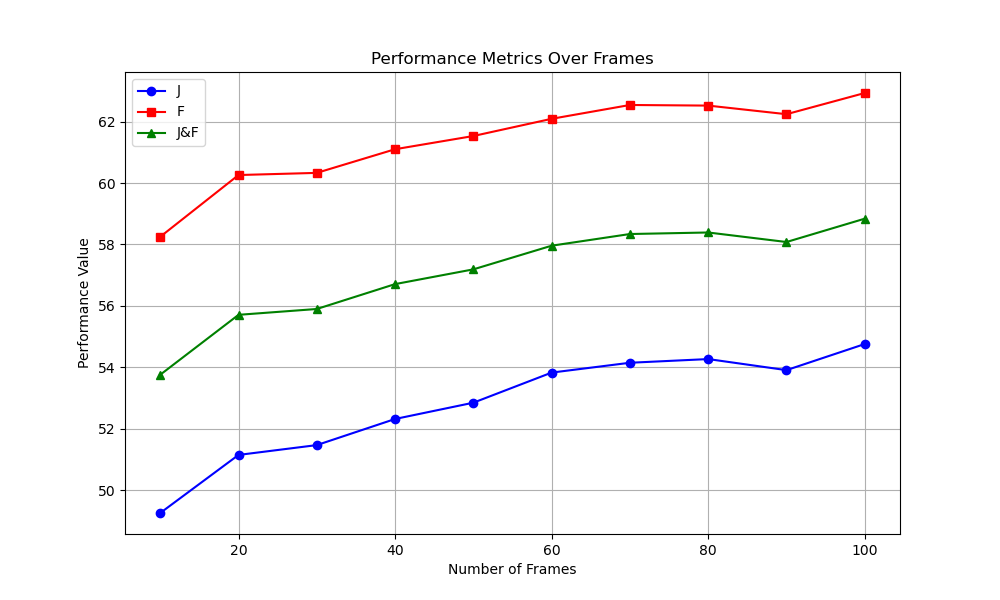}
    \vspace{-10pt}
    \caption{Scaling Curve of Frames}
    \label{fig:scaling_curve}
\end{figure}

\noindent\textbf{Scaling Curve of Frames.}
We further evaluate how the number of frames per video affects performance. Using the ASTC method with clips of 10 frames each, we increase the sampled frames from 10 to 100 in increments of 10, maintaining a compression ratio of 20\%. As illustrated in Fig.~\ref{fig:scaling_curve}, with the increasing number of frames, $J$, $F$ and $J\&F$ metrics consistently improve as the number of frames increases.
\vspace{5pt}

\noindent\textbf{<SEG> Token Granularity.}
We analyze the effect of varying the number of clips per \texttt{<SEG>} token in a video with 100 sampled frames, compressed via ASTC into 10 clips of 10 frames each. Specifically, we assign one \texttt{<SEG>} token per: (1) 10 clips (covering the entire video); (2) 5 clips (1 token per 20 frames); (3) 3 clips (approximately 1 token per 33 frames, with the last 4 clips sharing the same \texttt{<SEG>} token); (4) 2 clips (1 token per 50 frames); and (5) 1 clip (1 token per 10 frames). As shown in Table~\ref{tab:seg}, assigning one \texttt{<SEG>} token per clip yields the best performance across the evaluated metrics - $J$, $F$ and $J\&F$. 

\begin{table}[htbp]
\centering
\footnotesize
\begin{minipage}[t]{0.48\textwidth}
\centering
\begin{tabular}{l|ccc}
\toprule
Method & $J$ & $F$ & $J\&F$\\
\midrule
Average Pooling & 52.9 & 60.7 & 56.8  \\
Max Pooling & 53.3 & 61.5 & 57.4  \\
Token Prune & 51.1 & 59.5 & 55.3  \\
Token Merge & 52.8 & 60.7 & 56.8 \\
ASTC & 54.4 & 62.6 & 58.5 \\
\bottomrule
\end{tabular}
\vspace{5pt}
\caption{Compression Strategy Comparison}
\label{tab:compression}
\end{minipage}
\hfill
\begin{minipage}[t]{0.48\textwidth}
\centering
\begin{tabular}{c|ccc}
\toprule
Clips per Token & $J$ & $F$ & $J\&F$\\
\midrule
10 & 54.8 & 62.9 & 58.9  \\
5 & 54.7 & 63.1 & 58.9  \\
3 & 54.8 & 63.1 & 59.0 \\
2 & 54.9 & 63.3 & 59.1 \\
1 & 55.3 & 63.5 & 59.4 \\
\bottomrule
\end{tabular}
\vspace{5pt}
\caption{Effect of <SEG> Token Granularity}
\label{tab:seg}
\end{minipage}
\end{table}

\vspace*{-10pt}
\subsection{Limitations and Future Work}
Although our SVAC achieves SOTA performance, several limitations remain. First, our method may struggle with objects that appear briefly in the video sequence, as their temporal information might be lost during compression and attention computation. Second, challenging scenarios involving extreme occlusions or dramatic appearance changes can still pose difficulties, as the current number of segmentation tokens may not be sufficient to maintain consistent object tracking under such complex conditions.

Future work could address these limitations through several directions: (1) incorporating advanced motion prediction mechanisms to better handle brief object appearances and disappearances; (2) developing adaptive token allocation strategies that can dynamically adjust based on scene complexity; and (3) exploring multi-scale temporal feature fusion techniques to enhance the model's robustness to appearance variations. 

\section{Conclusion}
\label{sec:conclusion}
In this work, we propose SVAC, a novel framework for Referring Video Object Segmentation (RVOS) that effectively scales up both the number of processed video frames and segmentation tokens. We introduce Anchor-Based Spatio-Temporal Compression (ASTC), which processes more frames to better leverage MLLM’s potential by maintaining an anchor frame and a composite compressed representation of other frames for each clip, and Clip-Specific Allocation (CSA), which increases the number of segmentation tokens through clip-wise allocation to better capture motion patterns. Extensive experiments across multiple VOS benchmarks demonstrate that SVAC achieves state-of-the-art performance while maintaining computational efficiency. Our work provides a promising direction for advancing text-guided video understanding and segmentation tasks.

\bibliography{egbib}
\end{document}